\documentclass[twoside,11pt]{article}

\usepackage{blindtext}

\usepackage{booktabs}
\usepackage{subcaption}
\usepackage[preprint]{main}

\usepackage{lastpage}

\ShortHeadings{JaxARC}{Aadam et al.}
\firstpageno{1}

\begin{document}

\title{JaxARC: A High-Performance JAX-based Environment for Abstraction and Reasoning Research}

\author{\name Aadam \email mraadam@iu.edu \\
       \name Monu Verma \email monuverm@iu.edu \\
       \name Mohamed Abdel-Mottaleb \email mabdelm@iu.edu \\
       \addr Luddy School of Informatics, Computing, and Engineering\\
       Indiana University\\
       Indianapolis, IN 46202, USA}

\maketitle

\begin{abstract}
The Abstraction and Reasoning Corpus (ARC) tests AI systems' ability to perform human-like inductive reasoning from a few demonstration pairs. Existing Gymnasium-based RL environments severely limit experimental scale due to computational bottlenecks. We present JaxARC, an open-source, high-performance RL environment for ARC implemented in JAX. Its functional, stateless architecture enables massive parallelism, achieving 38-5,439× speedup over Gymnasium at matched batch sizes, with peak throughput of 790M steps/second. JaxARC supports multiple ARC datasets, flexible action spaces, composable wrappers, and configuration-driven reproducibility, enabling large-scale RL research previously computationally infeasible.
\end{abstract}


\section{Introduction}

The Abstraction and Reasoning Corpus (ARC) \citep{cholletMeasureIntelligence2019} challenges AI systems to solve novel reasoning puzzles by inferring transformation rules from a few examples; a capability central to human intelligence but largely absent in current state-of-the-art AI systems. Despite approaches ranging from program synthesis to large language models, ARC remains largely unsolved \citep{cholletARCPrize20242025}.

Reinforcement Learning (RL) offers a promising avenue for solving ARC, leveraging sequential decision-making to construct solutions iteratively, a capability distinct from one-shot synthesis methods and quite similar to how humans approach these problems. Despite this potential, RL research on ARC has been stifled by the lack of scalable simulation infrastructure. ARCLE \citep{leeARCLEAbstractionReasonin2024} established the first standardized Gymnasium-based RL environment for ARC, framing its challenges in RL terms: sparse rewards, vast combinatorial action spaces, and extreme diversity requiring strong generalization. However, ARCLE's Python-based, object-oriented architecture fundamentally limits scalability. Modern RL algorithms, like like PPO \citep{schulmanProximalPolicyOptimization2017} and meta-learning approaches \citep{beckTutorialMetaReinforcementLearning2025}, require billions of environment steps and massive parallelization for sample-efficient learning. Python's performance limitations create a critical bottelneck, where a single experiment might take days or weeks, preventing researchers from exploring and iterating on algorithmic innovations quickly \citep{hesselPodracerArchitecturesScalable2021}.

In this paper, we introduce \textbf{JaxARC}, a high-performance RL environment designed to overcome these scalability limitations. Built entirely in JAX \citep{jax2018github}, JaxARC leverages a functional, stateless architecture to enable automatic vectorization and XLA compilation on hardware accelerators (GPUs/TPUs). Unlike existing tools, JaxARC provides a unified, efficient interface for multiple ARC datasets and custom tasks; achieves massive throughput improvements demonstrating 38-5,439× speedups over ARCLE at matched batch sizes with peak rates of 790M steps/second; and integrates seamlessly with the growing JAX RL ecosystem, supporting composable wrappers and configuration-driven reproducibility. 

JaxARC is available at \url{https://github.com/aadimator/JaxARC} under the MIT license, with comprehensive documentation at \url{https://jaxarc.readthedocs.io}.

\section{JaxARC Architecture}

JaxARC's design prioritizes performance through functional purity while maintaining the familiar API conventions of Gymnasium-style environments \citep{towersGymnasiumStandardInterface2025}.

\textbf{Functional core:} JaxARC implements a purely functional API where \texttt{reset} and \texttt{step} are stateless pure functions that take explicit state and parameters rather than mutating internal state. All components are JAX pytrees, enabling JAX transformations (\texttt{jit}, \texttt{vmap}, \texttt{pmap}). The state representation uses fixed-size arrays with padding for variable-length grids, allowing efficient batching without dynamic shapes that would prevent JIT compilation. The API is built on top of the Stoa \citep{toledoStoaJAXnativeInterface2025} interface, facilitating interoperability with other JAX RL libraries.

\textbf{Task Management:} JaxARC's task buffer pre-loads and stacks all tasks into fixed-shape JAX arrays at initialization, converting variable-size ARC grids (1-30×1-30) into padded 30×30 arrays with mask tensors indicating valid regions. This eliminates runtime data loading and enables efficient sampling within JIT-compiled code through JAX's PRNG-based indexing. The system supports cross-split loading (train/evaluation), named subsets, lazy parsing to minimize memory for large datasets, and custom dataset formats through a parser interface.

\textbf{Flexible Action Spaces:} The core environment accepts an operation id plus a binary selection mask over the grid. Alternatively, different wrappers provide flexible selection mechansims: point actions $(row, col, op)$ yeild lowest action space size; bounding-box actions $(r_1,c_1,r_2,c_2,op)$ expand the action space considerably but offer richer region selection. All representations become masks before execution so algorithms can interchange them without changing downstream logic. Futhermore, a mechanism to select a subset of operations allows researchers to define smaller action spaces for specific tasks, reducing complexity.

\textbf{Composable wrappers:} Observation wrappers add channels (clipboard, original input grid, target grid, and $N$-contextual demo pairs) by concatenation along the channel axis. Action wrappers map point and bounding-box parameterizations to the internal mask form and optionally flatten composite discrete spaces for agent compatibility. All wrappers are pure and stackable such that composing several adds negligible overhead relative to the base environment.

\textbf{Configuration:} Hydra-based YAML configuration specifies datasets, action spaces, rewards, and wrappers. The \texttt{make()} function instantiates environments from identifiers or config objects, enabling reproducible experiments.

\textbf{Reward Structures:} JaxARC addresses ARC's sparse reward challenge through configurable reward shaping. The reward function combines four components: (1) \textit{similarity reward} (training only): shaped reward proportional to pixel-wise similarity improvement between working and target grids; (2) \textit{success bonus}: awarded only when agent submits with 100\% similarity; (3) \textit{step penalty}: small per-step cost encouraging efficiency; (4) \textit{unsolved submission penalty}: discourages premature submission. Mode-aware computation applies similarity shaping only during training, preventing target grid exploitation during evaluation. 

\section{Performance Evaluation}

We benchmarked JaxARC against ARCLE across three hardware platforms: Apple M2 Pro CPU (12 cores), NVIDIA RTX 3090 GPU (24GB), and NVIDIA H100 GPU (80GB). All benchmarks used a fixed MiniARC task with 100 steps per environment, averaged over multiple runs.

\paragraph{Configurations}
We evaluate two ARCLE baselines and two JaxARC variants. ARCLE: (1) \texttt{arcle-sync} (sequential \texttt{SyncVectorEnv}); (2) \texttt{arcle-async} (multiprocessing \texttt{AsyncVectorEnv}). JaxARC: (1) \texttt{jaxarc-jit} (single-device vectorization via \texttt{vmap}); (2) \texttt{jaxarc-pmap} (multi-device parallelization on H100). 

\paragraph{Results}

\begin{figure}[h]
\centering
\includegraphics[width=1\linewidth]{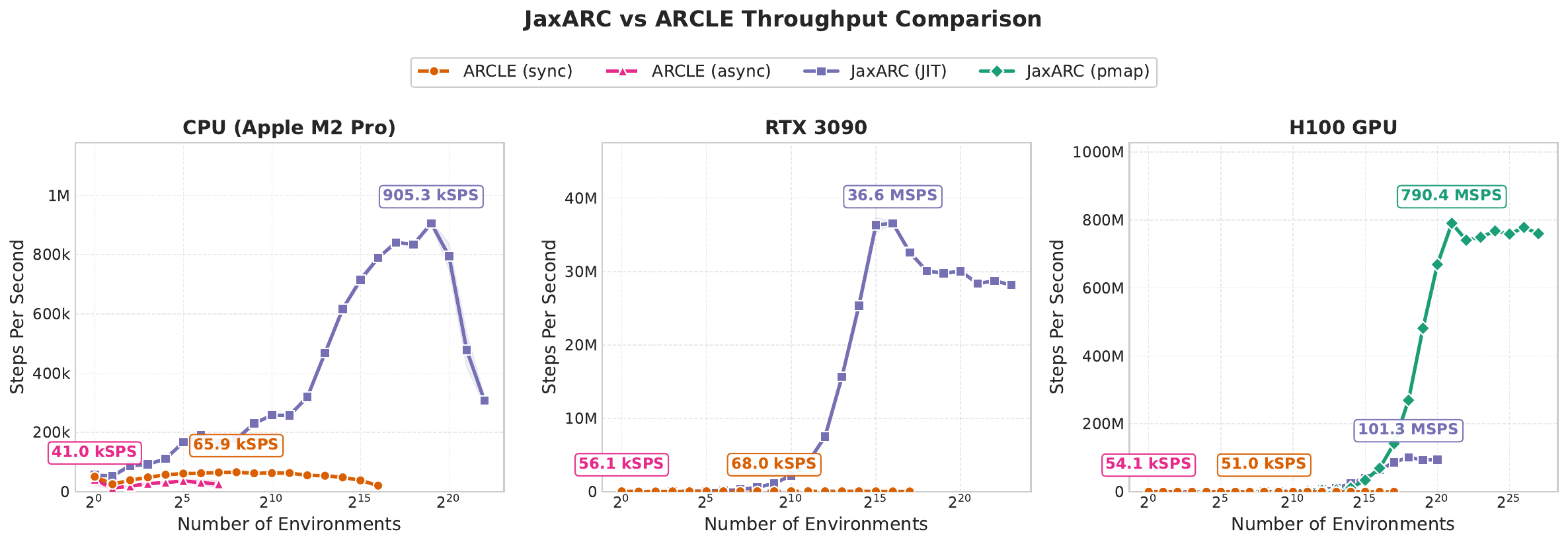}
\caption{Throughput (steps/second) vs number of parallel environments on CPU (left), RTX 3090 (center), and H100 (right). JaxARC scales near-linearly while ARCLE saturates due to Python/multiprocessing overhead.}
\label{fig:throughput}
\end{figure}

Figure~\ref{fig:throughput} shows throughput scaling across batch sizes. JaxARC demonstrates near-linear scaling on all platforms, while ARCLE saturates quickly. Table~\ref{tab:performance} compares best-case performance at matched batch sizes where both frameworks have data.

\begin{table}[t]
\centering
\caption{Peak throughput at matched batch sizes. Best speedup achieved when comparing identical batch sizes where both frameworks have valid measurements.}
\label{tab:performance}
\begin{tabular}{@{}lrrrr@{}}
\toprule
\textbf{Platform} & \textbf{Batch Size} & \textbf{JaxARC} & \textbf{ARCLE} & \textbf{Speedup} \\
\midrule
CPU (M2 Pro) & 65,536 & 789K & 21K & 38× \\
RTX 3090 & 131,072 & 32.6M & 36K & 903× \\
H100 GPU & 131,072 & 142M & 26K & 5,439× \\
\bottomrule
\end{tabular}
\end{table}

At matched batch sizes, JaxARC achieves 38-5,439× speedup (Table~\ref{tab:performance}). Speedups are modest (1-10×) at small scales (1-256 envs) where Python overhead is amortized, but grow dramatically at large scales where ARCLE's per-step Python interpreter costs and IPC overhead dominate. JaxARC reaches 790M steps/sec on H100 at 2M environments, completing $10^9$ steps in 1.3s. At higher batch sizes, ARCLE configurations consistently crashed the system, resulting in missing data points beyond 131K environments.  Comparing best throughputs across all batch sizes yields up to 14,000× speedup, though this is not a strictly matched comparison since JaxARC operates at scales ARCLE cannot reach.

\paragraph{Impact} This performance enables a shift in the experimental paradigms, reducing experiment time from weeks to hours. JaxARC enables population-based training \citep{langeEvosaxJAXbasedEvolution2022}, architecture search, and meta-learning algorithms like MAML \citep{finnModelagnosticMetalearningFast2017} requiring millions of episodes. Additionally, JaxARC's deterministic execution from PRNG seeds enables perfect reproducibility across hardware platforms. The same configuration file and seed produce identical rollouts on CPU, GPU, or TPU, eliminating hardware-dependent variance that confounds experimental results in traditional RL frameworks.

\section{Related Work}

\textbf{JAX RL libraries:} Gymnax \citep{gymnax2022github} provides JAX implementations of classic RL benchmarks; Jumanji \citep{bonnetJumanjiDiverseSuite2023} offers combinatorial optimization tasks; Brax \citep{freemanBraxDifferentiablePhysics2021} focuses on continuous control. JaxARC fills the gap for abstract reasoning benchmarks, complementing these libraries.

\textbf{ARC challenge:} François Chollet's original benchmark \citep{cholletMeasureIntelligence2019} sparked numerous approaches: program synthesis \citep{ainoosonNeurodiversityInspiredSolverAbstraction2023}, neuro-symbolic methods \citep{buttCodeItSelfimprovingLanguage2024,alfordNeuralGuidedBidirectionalProgram2022}, and language models \citep{camposampieroAbstractVisualReasoning2023,leeReasoningAbilitiesLarge2025}. The recent ARC Prize \citep{cholletARCPrize20242025} renewed interest with a \$1M prize, highlighting the need for scalable experimentation tools.

\textbf{ARC environments:} ARCLE \citep{leeARCLEAbstractionReasonin2024} pioneered RL for ARC with a Gymnasium interface but suffers from Python performance limitations. 

\section{Conclusion}

JaxARC provides a high-performance environment for RL research on abstract reasoning, achieving tremendous speedup over Python implementations. This positions ARC as a benchmark suitable for modern deep RL, meta-learning, and large-scale architecture search. Future work includes additional datasets, baseline agents, and novel algorithmic development on top of this platform to solve abstract reasoning using the sequential decision making paradigm. By removing the computational barrier, JaxARC enables focus on fundamental algorithmic challenges of abstract reasoning.

\acks{We thank François Chollet for creating ARC, the ARCLE developers for pioneering RL approaches to ARC, and Edan Toledo for his work on Stoa API design. 
This research received no external funding. The authors declare no competing interests.}

\bibliography{references}

\newpage

\appendix

\section{System Architecture}
\label{app:architecture}

JaxARC's architecture is designed for high performance and flexibility. As shown in Figure~\ref{fig:architecture}, the system consists of five main layers:

\begin{figure}[h]
    \centering
    \includegraphics[width=0.9\linewidth]{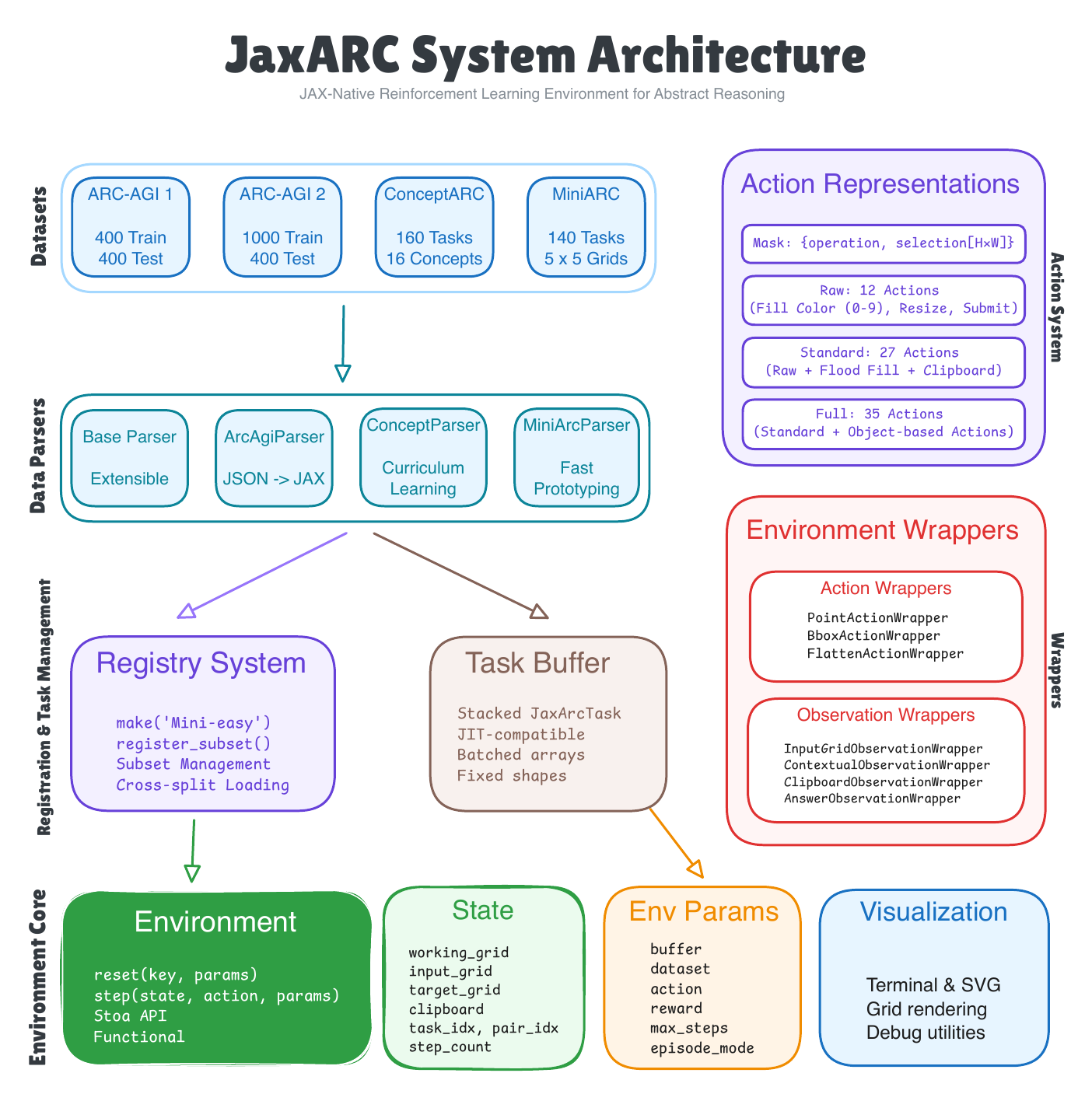}
    \caption{JaxARC System Architecture. The functional core enables seamless JAX transformations, while the task buffer ensures efficient data loading.}
    \label{fig:architecture}
\end{figure}

\begin{itemize}
    \item \textbf{Parsers:} Format-agnostic readers that enable support for custom datasets.
    \item \textbf{Registry:} Handles task management, subset definitions, and dataset downloads.
    \item \textbf{Buffer:} Manages memory and enables efficient sampling by pre-stacking tasks into fixed-size arrays.
    \item \textbf{Core:} Implements the functional transition logic using the Stoa API, ensuring statelessness and JAX compatibility.
    \item \textbf{Wrappers:} Provide composable transformations for observation and action spaces, allowing researchers to adapt the environment to various RL paradigms.
\end{itemize}

\section{Visualization and Tasks}
\label{app:visualization}

\begin{figure}[h]
    \centering
    \begin{minipage}[c]{0.23\linewidth}
        \begin{subfigure}[b]{1.0\linewidth}
            \centering
            \includegraphics[width=\linewidth]{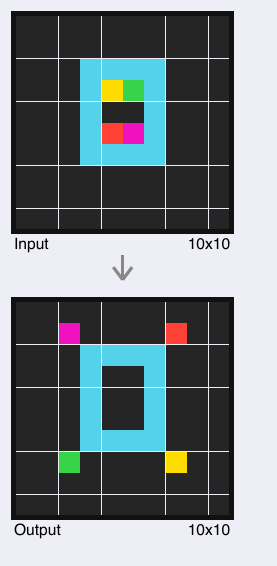}
            \caption{I/O Pair}
            \label{fig:viz_pair}
        \end{subfigure}
    \end{minipage}
    \hfill
    \begin{minipage}[c]{0.74\linewidth}
        \begin{subfigure}[b]{0.32\linewidth}
            \centering
            \includegraphics[width=\linewidth]{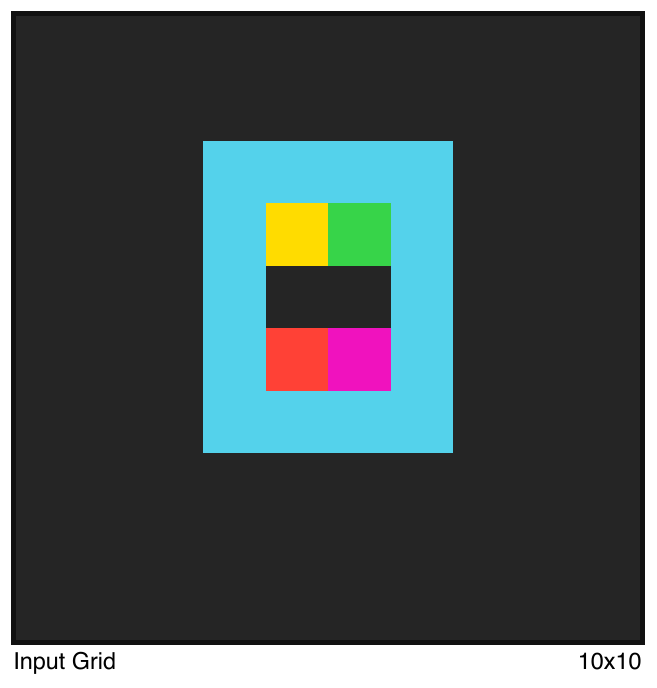}
            \caption{Single Grid}
            \label{fig:viz_single}
        \end{subfigure}
        \hfill
        \begin{subfigure}[b]{0.64\linewidth}
            \centering
            \includegraphics[width=\linewidth]{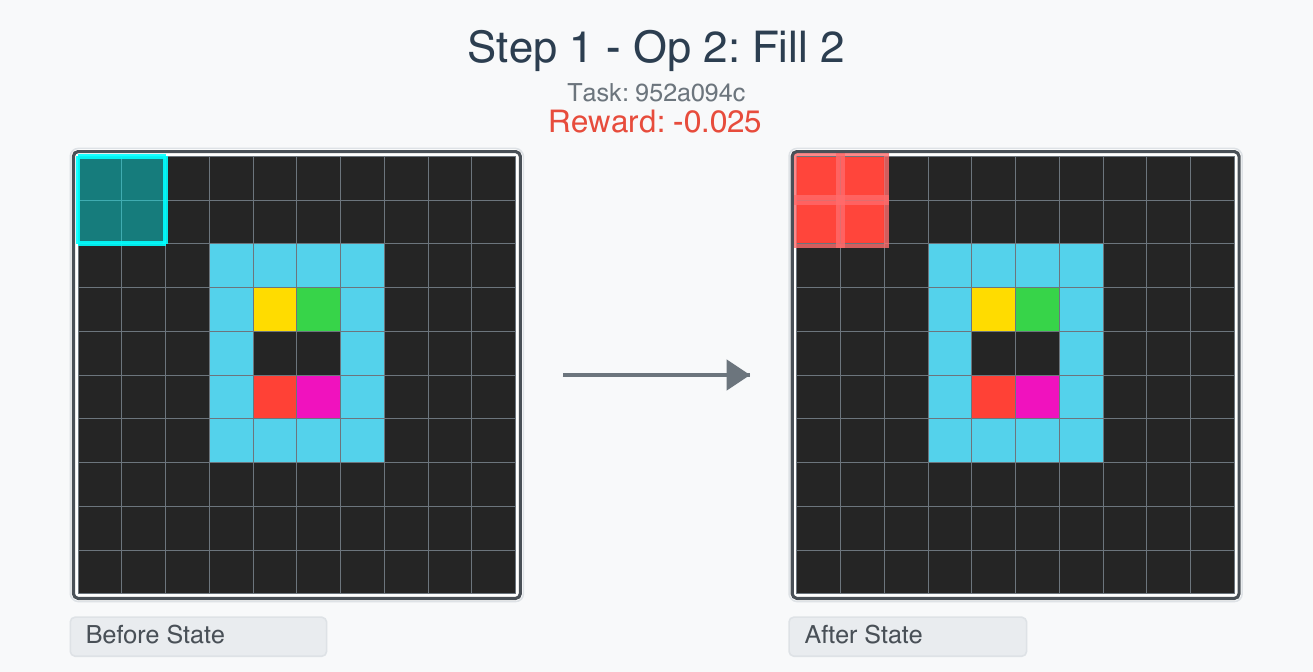}
            \caption{RL Step}
            \label{fig:viz_rl}
        \end{subfigure}
        
        \vspace{0.5em}
        
        \begin{subfigure}[b]{1.0\linewidth}
            \centering
            \includegraphics[width=\linewidth]{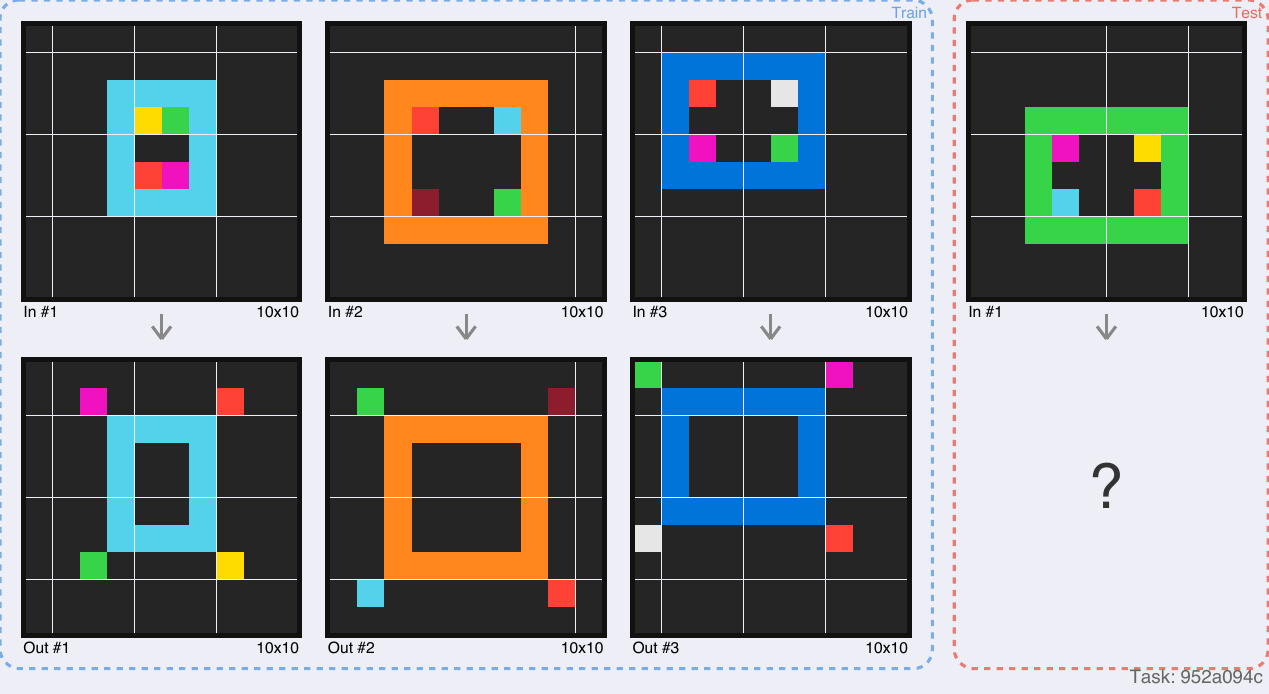}
            \caption{Complete Task View}
            \label{fig:viz_complete}
        \end{subfigure}
    \end{minipage}
    
    \caption{JaxARC Visualization Modes. (a) Input/Output pair view showing transformation. (b) Single grid view. (c) RL step visualization showing agent interaction. (d) Complete task view showing all training pairs and test input.}
    \label{fig:viz_modes}
\end{figure}

JaxARC supports rich visualization capabilities, including terminal-based rendering and SVG output, which are essential for understanding agent behavior and debugging complex reasoning tasks. Figure~\ref{fig:viz_modes} illustrates the four primary visualization modes:

\begin{itemize}
    \item \textbf{I/O Pair (a):} Displays a single input-output demonstration pair side-by-side. This view is fundamental for identifying the specific transformation rule required for a sub-problem.
    \item \textbf{Single Grid (b):} Focuses on a specific grid state (input, output, or intermediate). This is useful for detailed pixel-level inspection of the agent's current workspace.
    \item \textbf{RL Step (c):} Visualizes a single transition in the environment, showing the state before and after an action, along with the action taken (highlighted) and the immediate reward received. This mode is crucial for debugging RL policies and verifying reward functions.
    \item \textbf{Complete Task View (d):} Provides a comprehensive overview of the entire ARC task, displaying all training demonstration pairs (inputs and outputs) and the test input. This view represents the full context available to the agent for few-shot inference.
\end{itemize}

These visualization tools allow researchers to qualitatively assess agent performance and diagnose failure modes effectively.

\section{Usage Example}
\label{app:usage}

The following code demonstrates basic JaxARC usage, including environment creation, single-env interaction, vectorization, and JIT compilation:

\begin{verbatim}
import jax
import jax.numpy as jnp
from jaxarc import make

# Create environment with specific task
env, env_params = make("Mini-Most_Common_color_l6ab0lf3xztbyxsu3p", auto_download=True)

# Single environment interaction
key = jax.random.PRNGKey(0)
state, timestep = env.reset(key, env_params=env_params)

# Sample action and step
action_space = env.action_space(env_params)
key, subkey = jax.random.split(key)
action = action_space.sample(subkey)
next_state, next_timestep = env.step(state, action, env_params=env_params)
print(f"Reward: {next_timestep.reward}")

# Vectorize across 1024 environments
num_envs = 1024
key, *reset_keys = jax.random.split(key, num_envs + 1)
reset_keys = jnp.array(reset_keys)

# vmap over reset keys, broadcast env_params
vmapped_reset = jax.vmap(env.reset, in_axes=(0, None))
states, timesteps = vmapped_reset(reset_keys, env_params)

# JIT-compiled rollout for performance
@jax.jit
def rollout_step(state, key):
    action = action_space.sample(key)
    return env.step(state, action, env_params=env_params)

# Execute vectorized rollout (1024 envs in parallel)
key, *step_keys = jax.random.split(key, num_envs + 1)
step_keys = jnp.array(step_keys)
vmapped_step = jax.vmap(rollout_step)
next_states, next_timesteps = vmapped_step(states, step_keys)
# Executes on GPU/TPU with no code changes
\end{verbatim}

\section{Benchmark Details}
\label{app:benchmark}

All benchmarks used MiniARC task \texttt{Most\_Common\_color\_l6ab0lf3xztbyxsu3p} with consistent experimental setup across three hardware platforms.

\textbf{Hardware:} CPU: Apple M2 Pro (12-core, 16GB RAM); RTX 3090: NVIDIA GeForce RTX 3090 (24GB VRAM); H100: NVIDIA H100 GPU (80GB HBM3, single device).

\textbf{Software:} Python 3.13.5, JAX 0.6.2, NumPy 2.3.1, Gymnasium 1.2.0. ARCLE environment via \texttt{ARCLE/RawARCEnv-v0} with \texttt{MiniARCLoader} and $5\times5$ max grid size. Both frameworks used bounding-box action wrappers for consistency.

\textbf{Methodology:} Fixed 100 steps per environment with auto-reset on episode termination. Batch sizes varied from $2^0$ to $2^{20}$ (1 to 1,048,576 environments). Each configuration repeated 3--10 times using \texttt{timeit.repeat}. JaxARC measurements separate JIT compilation time (one-time cost) from runtime; ARCLE has no compilation overhead. Throughput computed as total steps (batch size $\times$ 100) divided by execution time, excluding compilation.

\textbf{ARCLE configurations:} (1) \texttt{arcle-sync}: Gymnasium's \texttt{SyncVectorEnv} executing environments sequentially in a single process; (2) \texttt{arcle-async}: \texttt{AsyncVectorEnv} distributing environments across OS-level worker processes for parallel execution.

\textbf{JaxARC configurations:} (1) \texttt{jaxarc-jit}: Single-device vectorization via \texttt{jax.vmap} with JIT compilation; (2) \texttt{jaxarc-pmap}: Multi-device parallelization via \texttt{jax.pmap} (H100 only, tested but single-device results reported here). All JaxARC runs used scan unroll factor of 1.

\textbf{Matched comparison:} Table~\ref{tab:performance} reports maximum speedup at batch sizes where both frameworks successfully ran. ARCLE configurations crashed beyond approximately 131,072 environments due to memory exhaustion from process duplication in vectorized environments. JaxARC scaled to 2,097,152 environments (2M), limited only by available device memory. At small batch sizes (1--256 envs), Python overhead dominates for both frameworks; at large scales ($>$65K envs), JaxARC's XLA-compiled kernels and on-device execution eliminate host-device transfer bottlenecks, yielding dramatic performance advantages.

\subsection{Log-Scale Throughput Analysis}
\begin{figure}[h]
    \centering
    \includegraphics[width=1\linewidth]{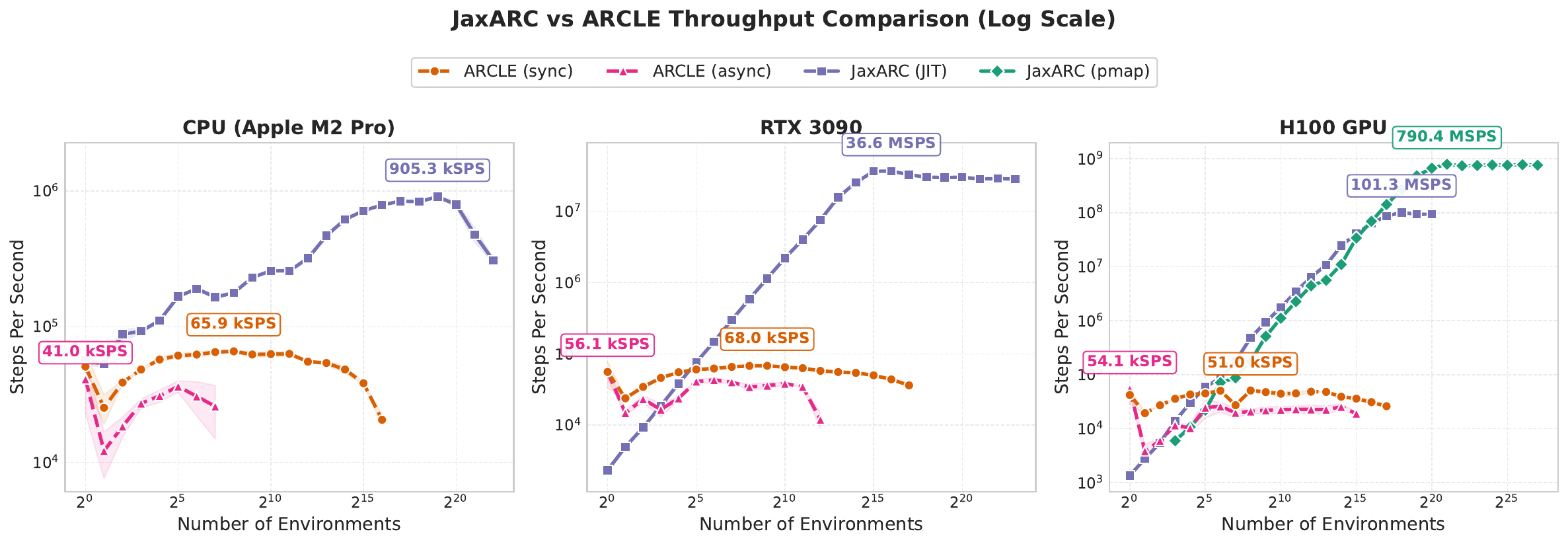}
    \caption{Throughput (steps/second) vs number of parallel environments on Log Scale for CPU (left), RTX 3090 (center), and H100 (right).}
    \label{fig:throughput_log}
\end{figure}

Figure~\ref{fig:throughput_log} presents the throughput scaling on a logarithmic scale, highlighting the performance characteristics across the entire range of batch sizes.

The log-log plots reveal three distinct performance regimes driven by the interplay between fixed overheads and parallelism:

\begin{itemize}
    \item \textbf{Latency-Dominated Regime ($N < 32$):} On accelerators, ARCLE initially outperforms JaxARC. At this scale, the fixed overhead of JAX kernel dispatch and host-device data transfer dominates execution time. ARCLE's lightweight Python loop incurs lower latency for trivial batch sizes, making it competitive for debugging or single-agent interaction.
    
    \item \textbf{Linear Scaling Regime ($32 \le N < 10^5$):} As batch size increases, JaxARC amortizes its dispatch overhead and enters a phase of near-perfect linear scaling. Throughput doubles with every doubling of environments. The crossover point occurs consistently around $N=32$, after which JaxARC rapidly outpaces ARCLE. Conversely, ARCLE's throughput remains flat, indicating it is entirely CPU-bound and unable to leverage parallelism.
    
    \item \textbf{Saturation Regime ($N \ge 10^5$):} JaxARC continues to scale until hardware limits are reached. On the RTX 3090, performance plateaus around $N=2^{16}$ (65k) at 36.6 MSPS. On the H100, scaling continues impressively up to $N=2^{21}$ (2M), peaking at 790.4 MSPS. The CPU curve shows a similar peak at $N=2^{19}$ before degrading due to cache thrashing.
\end{itemize}

These results highlight a critical trade-off: while standard Python environments offer lower latency for single-instance usage, they are fundamentally unsuited for high-throughput learning. JaxARC's functional architecture successfully unlocks the massive parallelism of modern hardware, providing the orders-of-magnitude speedup required for large-scale RL.

\subsection{Detailed Speedup Analysis}
Tables~\ref{tab:speedup_cpu_m2_pro}, \ref{tab:speedup_rtx_3090}, and \ref{tab:speedup_h100_gpu} provide a detailed breakdown of throughput and speedup at each common batch size where both JaxARC and ARCLE could be evaluated. The table only shows the highest-performing ARCLE configuration (either \texttt{arcle-sync} or \texttt{arcle-async}) or the JaxARC configuration (\texttt{jaxarc-jit} or \texttt{jaxarc-pmap}) at each batch size. This is to highlight the maximum achievable speedup at each scale.

\begin{itemize}
    \item \textbf{CPU Dominance:} On the M2 Pro CPU (Table~\ref{tab:speedup_cpu_m2_pro}), JaxARC is consistently faster than ARCLE across all batch sizes, with the gap widening significantly from 1.1× at $N=1$ to 38.2× at $N=65,536$.
    \item \textbf{Accelerator Crossover:} On GPU platforms (Tables~\ref{tab:speedup_rtx_3090} and \ref{tab:speedup_h100_gpu}), a distinct crossover occurs at $N=32$. Below this threshold, ARCLE's lower latency results in higher throughput. Above $N=32$, JaxARC's throughput explodes, while ARCLE stagnates.
    \item \textbf{Magnitude of Speedup:} The speedup factor is not constant but grows linearly with batch size. At $N=131,072$, JaxARC achieves a 903× speedup on the RTX 3090 and a staggering 5,439× speedup on the H100.
\end{itemize}

These tables demonstrate that while ARCLE is sufficient for small-scale debugging, it hits a hard performance ceiling around 60k SPS regardless of hardware. JaxARC removes this ceiling, allowing throughput to scale with the raw compute capability of the hardware, unlocking the potential for training on billions of frames.

\begin{table}[h]
\centering
\caption{Detailed speedup comparison on CPU (M2 Pro)}
\label{tab:speedup_cpu_m2_pro}
\begin{tabular}{lrrrrr}
\toprule
Batch Size & JaxARC SPS & Mode & ARCLE SPS & Mode & Speedup \\
\midrule
1 & 56,266 & jaxarc-jit & 50,823 & arcle-sync & 1.1× \\
2 & 53,035 & jaxarc-jit & 25,299 & arcle-sync & 2.1× \\
4 & 87,928 & jaxarc-jit & 38,816 & arcle-sync & 2.3× \\
8 & 92,525 & jaxarc-jit & 48,384 & arcle-sync & 1.9× \\
16 & 111,057 & jaxarc-jit & 57,336 & arcle-sync & 1.9× \\
32 & 166,687 & jaxarc-jit & 61,319 & arcle-sync & 2.7× \\
64 & 190,381 & jaxarc-jit & 62,324 & arcle-sync & 3.1× \\
128 & 165,107 & jaxarc-jit & 65,020 & arcle-sync & 2.5× \\
256 & 177,682 & jaxarc-jit & 65,871 & arcle-sync & 2.7× \\
512 & 230,453 & jaxarc-jit & 62,397 & arcle-sync & 3.7× \\
1,024 & 257,889 & jaxarc-jit & 62,820 & arcle-sync & 4.1× \\
2,048 & 257,877 & jaxarc-jit & 63,111 & arcle-sync & 4.1× \\
4,096 & 319,194 & jaxarc-jit & 55,270 & arcle-sync & 5.8× \\
8,192 & 467,960 & jaxarc-jit & 54,033 & arcle-sync & 8.7× \\
16,384 & 616,683 & jaxarc-jit & 48,383 & arcle-sync & 12.7× \\
32,768 & 714,731 & jaxarc-jit & 38,334 & arcle-sync & 18.6× \\
65,536 & 788,855 & jaxarc-jit & 20,634 & arcle-sync & 38.2× \\
\bottomrule
\end{tabular}
\end{table}

\begin{table}[h]
\centering
\caption{Detailed speedup comparison on RTX 3090}
\label{tab:speedup_rtx_3090}
\begin{tabular}{lrrrrr}
\toprule
Batch Size & JaxARC SPS & Mode & ARCLE SPS & Mode & Speedup \\
\midrule
1 & 2,327 & jaxarc-jit & 56,066 & arcle-async & 0.0× \\
2 & 4,948 & jaxarc-jit & 23,845 & arcle-sync & 0.2× \\
4 & 9,282 & jaxarc-jit & 34,396 & arcle-sync & 0.3× \\
8 & 18,667 & jaxarc-jit & 45,991 & arcle-sync & 0.4× \\
16 & 38,305 & jaxarc-jit & 55,162 & arcle-sync & 0.7× \\
32 & 76,158 & jaxarc-jit & 60,394 & arcle-sync & 1.3× \\
64 & 147,835 & jaxarc-jit & 62,266 & arcle-sync & 2.4× \\
128 & 299,201 & jaxarc-jit & 65,513 & arcle-sync & 4.6× \\
256 & 589,248 & jaxarc-jit & 67,777 & arcle-sync & 8.7× \\
512 & 1,141,716 & jaxarc-jit & 68,011 & arcle-sync & 16.8× \\
1,024 & 2,205,416 & jaxarc-jit & 65,038 & arcle-sync & 33.9× \\
2,048 & 4,010,852 & jaxarc-jit & 63,005 & arcle-sync & 63.7× \\
4,096 & 7,519,602 & jaxarc-jit & 57,698 & arcle-sync & 130.3× \\
8,192 & 15,672,773 & jaxarc-jit & 55,584 & arcle-sync & 282.0× \\
16,384 & 25,365,090 & jaxarc-jit & 54,126 & arcle-sync & 468.6× \\
32,768 & 36,337,555 & jaxarc-jit & 50,002 & arcle-sync & 726.7× \\
65,536 & 36,570,507 & jaxarc-jit & 43,765 & arcle-sync & 835.6× \\
131,072 & 32,613,474 & jaxarc-jit & 36,131 & arcle-sync & 902.6× \\
\bottomrule
\end{tabular}
\end{table}

\begin{table}[h]
\centering
\caption{Detailed speedup comparison on H100 GPU}
\label{tab:speedup_h100_gpu}
\begin{tabular}{lrrrrr}
\toprule
Batch Size & JaxARC SPS & Mode & ARCLE SPS & Mode & Speedup \\
\midrule
1 & 1,359 & jaxarc-jit & 54,082 & arcle-async & 0.0× \\
2 & 2,776 & jaxarc-jit & 19,375 & arcle-sync & 0.1× \\
4 & 5,482 & jaxarc-jit & 27,123 & arcle-sync & 0.2× \\
8 & 13,912 & jaxarc-jit & 35,775 & arcle-sync & 0.4× \\
16 & 30,045 & jaxarc-jit & 43,116 & arcle-sync & 0.7× \\
32 & 59,176 & jaxarc-jit & 44,860 & arcle-sync & 1.3× \\
64 & 86,458 & jaxarc-jit & 50,575 & arcle-sync & 1.7× \\
128 & 176,671 & jaxarc-jit & 27,236 & arcle-sync & 6.5× \\
256 & 482,233 & jaxarc-jit & 51,035 & arcle-sync & 9.4× \\
512 & 942,287 & jaxarc-jit & 47,571 & arcle-sync & 19.8× \\
1,024 & 1,766,400 & jaxarc-jit & 44,436 & arcle-sync & 39.8× \\
2,048 & 3,448,222 & jaxarc-jit & 44,965 & arcle-sync & 76.7× \\
4,096 & 6,440,953 & jaxarc-jit & 48,389 & arcle-sync & 133.1× \\
8,192 & 10,743,903 & jaxarc-jit & 47,972 & arcle-sync & 224.0× \\
16,384 & 24,598,104 & jaxarc-jit & 39,065 & arcle-sync & 629.7× \\
32,768 & 40,676,131 & jaxarc-jit & 36,002 & arcle-sync & 1129.8× \\
65,536 & 69,422,302 & jaxarc-pmap & 31,038 & arcle-sync & 2236.7× \\
131,072 & 141,892,015 & jaxarc-pmap & 26,089 & arcle-sync & 5438.8× \\
\bottomrule
\end{tabular}
\end{table}

\section{Action System Design}
\label{app:actions}

JaxARC implements a mask-based action system with 35 operations (IDs 0--34) organized into six categories: fill (0--9), flood fill (10--19), movement (20--23), transformation (24--27), editing (28--30), and special operations (31--34). Each action consists of an operation ID and a binary selection mask specifying which grid cells the operation affects.

\subsection{Design Rationale}

JaxARC's action system design is largely similar to ARCLE's, but it differs from it in a couple of key aspects driven by JAX compatibility and practical considerations:

\textbf{Unified mask-based selection:} Unlike ARCLE's dual-mode system with separate object state tracking (selected pixels, background pixels, rotation parity), JaxARC uses a single selection mask applied uniformly across all operations. This eliminates stateful mode switching incompatible with JAX's functional paradigm while simplifying the action space. When no selection is provided, operations auto-select the entire working grid area defined by \texttt{working\_grid\_mask}.

\textbf{BrainGridGame-compatible operations:} Object-oriented operations (move, rotate, flip) follow the interface of \url{https://braingridgame.com}, which collects human solution trajectories in the ARC-Interactive-History-Dataset\footnote{\url{https://github.com/neoneye/ARC-Interactive-History-Dataset}}. This design choice enables future imitation learning and behavioral cloning research using recorded human demonstrations. Operations work on bounding boxes extracted from selections: movement wraps within bounding box boundaries; rotation requires square bounding boxes; flipping applies to the full bounding box region.

\subsection{Operation Categories}

\textbf{Fill operations (0--9):} Set selected cells to colors 0--9. Applied element-wise to selection mask.

\textbf{Flood fill operations (10--19):} Flood fill from a single selected cell with colors 0--9. Requires exactly one selected cell; uses fixed-iteration algorithm (64 steps) for JAX compatibility.

\textbf{Movement operations (20--23):} Move selected region up/down/left/right with wrapping within bounding box. Auto-selects entire grid if no selection provided.

\textbf{Transformation operations (24--27):} Rotate selected region 90° clockwise or counterclockwise (requires square bounding box) or flip horizontally/vertically within bounding box.

\textbf{Editing operations (28--30):} Copy selection to clipboard, paste clipboard to selection, or cut selection to clipboard (copy then clear).

\textbf{Special operations (31--34):} Clear selected cells (31), copy input grid to working grid (32), resize grid to selection bounding box (33), submit solution and terminate episode (34).

\subsection{Comparison with ARCLE}

ARCLE's O2ARCEnv provides richer object manipulation through explicit object state (\texttt{selected}, \texttt{active}, \texttt{object}, \texttt{object\_sel}) and stateful operations tracking rotation parity and background pixels. While this enables more nuanced object editing, it introduces mutable state incompatible with JAX transformations and increases action space complexity. JaxARC trades fine-grained object state for functional purity and simplicity. 

Both frameworks support trajectory collection for imitation learning, though ARCLE's official trajectory dataset (ARCTraj) uses their object-oriented interface while JaxARC trajectories align with BrainGridGame's simpler interaction model.

\section{Stoix Integration}
\label{app:stoix}

JaxARC integrates seamlessly with Stoix \citep{toledoStoixDistributedSingleagent2024}, a JAX-native RL library, as it adheres to the Stoa API standard, which provides a unified interface for JAX environments. This compliance enables JaxARC to work with multiple RL algorithms already implemented in the Stoix codebase without modification. We demonstrate this integration through \texttt{jaxarc-baselines}\footnote{\url{https://github.com/aadimator/jaxarc-baselines}}, a repository providing training scripts, configuration files, and experimental infrastructure for running Stoix algorithms on ARC tasks.

\subsection{Implementation via Monkey-Patching}

Rather than forking Stoix or requiring users to install modified versions, \texttt{jaxarc-baselines} uses lightweight monkey-patching to inject JaxARC support into Stoix's environment discovery system:

\begin{verbatim}
# From jaxarc-baselines/run_experiment.py
from stoix.utils import make_env as stoix_make_env_module
stoix_make_env_module.make = get_custom_make_fn(stoix_make_env_module.make)
\end{verbatim}

The wrapper function checks if \texttt{config.env.env\_name == "jaxarc"}; if true, it instantiates JaxARC with appropriate observation/action wrappers and vectorization. Otherwise, it delegates to Stoix's original factory, preserving compatibility with Gymnax, Jumanji, and other environments. Similarly, we can extend Stoix's logger to track ARC-specific metrics (success rate, steps to solve, similarity progression) using the same monkey-patching approach. This approach requires no changes to Stoix's codebase. Users can simply utilize the \texttt{run\_experiment.py} script from \texttt{jaxarc-baselines} instead of Stoix's default entry point.

\subsection{Configuration and Reproducibility}

The \texttt{jaxarc-baselines} repository provides Hydra configuration files specifying environment parameters (task subsets, reward shaping, observation wrappers), network architectures, and algorithm hyperparameters. 

All configuration files are version-controlled and can be easily modified for ablation studies. Scripts for launching SLURM-based hyperparameter sweeps are also included, enabling large-scale experiments across observation configurations, network architectures, and task subsets.

The integration demonstrates that JaxARC is not a standalone, isolated environment but a first-class member of the JAX RL ecosystem. Researchers can immediately leverage a wide range of RL algorithm implementations from Stoix without writing glue code. By providing reproducible baselines with open configurations, we lower the barrier to entry for researchers interested in abstract reasoning tasks.

\section{PPO Results: Observation Configuration Ablation}
\label{app:ppo-results}

We conducted experiments to evaluate how different observation configurations affect PPO performance on MiniARC tasks. This ablation study tests whether providing additional context, such as answer grids, input grids, or demonstration examples, helps or hinders CNN-based policy learning.

\subsection{Experimental Setup}

\textbf{Algorithm:} Stoix FF-PPO (feed-forward PPO) with the Anakin architecture (1,024 parallel environments). Hyperparameters: actor/critic learning rates $3 \times 10^{-4}$ with linear decay, rollout length 128, 4 PPO epochs over 16 minibatches per update, $\gamma=0.99$, GAE $\lambda=0.95$, clip coefficient $\varepsilon=0.2$, entropy coefficient 0.01, value loss coefficient 0.5, max gradient norm 0.5. Training ran for 500M total environment steps.

\textbf{Network architecture:} Shallow CNN with two $3\times3$ convolutional layers (32 feature channels, SiLU activation) followed by two fully-connected layers (128 units each). The CNN processes multi-channel grid observations; input channel count varies by observation wrapper configuration (1--13 channels).

\textbf{Environment configuration:} The dataset selected is the MiniARC dataset (149 tasks) with point-action interface (discrete actions flattened to single discrete space), 150-step episode limit, balanced reward shaping (success bonus +10.0, step penalty $-0.02$, unsolved submission penalty $-1.0$, similarity weight 1.0). Auto-reset enabled for seamless episode transitions within vectorized rollouts.

\textbf{Observation wrappers:} JaxARC provides composable wrappers that augment the base observation (working grid, 1 channel) by adding channels:
\begin{enumerate}
\item \textbf{AnswerObservationWrapper}: Adds the target output grid as one additional channel (+1), providing the desired solution as a ``hint''.
\item \textbf{InputGridObservationWrapper}: Adds the original input grid as one additional channel (+1), maintaining reference to the initial state.
\item \textbf{ContextualObservationWrapper}: Adds demonstration pairs from the task as additional channels (+10 for default 5 pairs: each pair contributes 2 channels for input and output grids). During training, excludes the current pair being solved; during testing, includes all available demonstration pairs.
\end{enumerate}

We tested seven wrapper configurations: (1) \textit{None}: working grid only (1 ch); (2) \textit{Answer}: + answer wrapper (2 ch); (3) \textit{Input}: + input wrapper (2 ch); (4) \textit{Contextual}: + contextual wrapper (11 ch); (5) \textit{Input + Answer}: + both wrappers (3 ch); (6) \textit{Input + Contextual}: + both wrappers (12 ch); (7) \textit{Input + Contextual + Answer}: all three wrappers (13 ch). Each configuration ran for 5 random seeds.

\textbf{Evaluation:} Success rate measured as percentage of episodes achieving 100\% pixel-wise similarity between working grid and target upon submission. Throughput (steps per second) measured on a single H100 GPU.

\subsection{Results and Analysis}

Figure~\ref{fig:ppo-obs-ablation} shows success rate and throughput curves for seven observation configurations. The most informative configurations are:

\begin{figure}[t]
\centering
\includegraphics[width=0.48\linewidth]{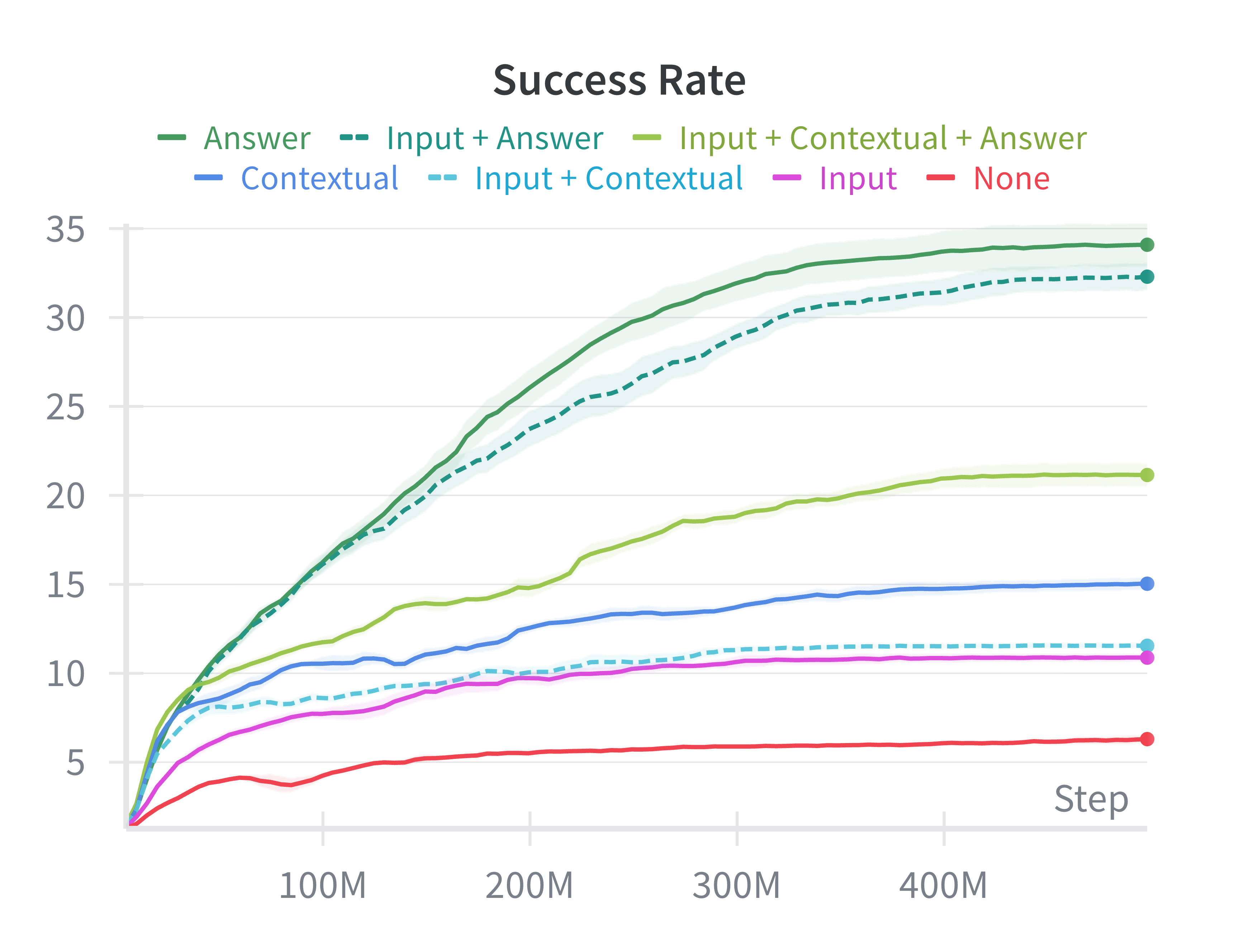}
\hfill
\includegraphics[width=0.48\linewidth]{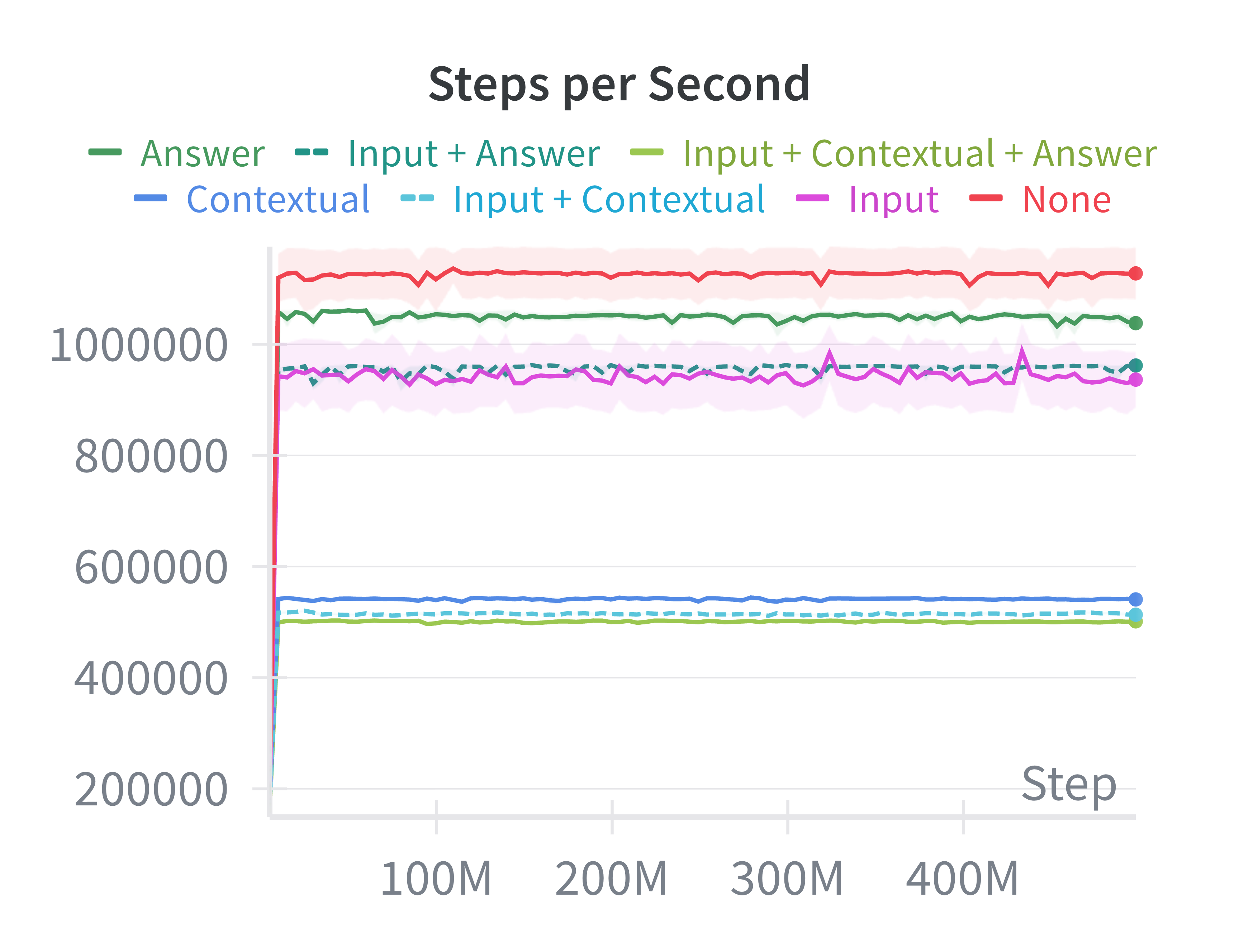}
\caption{Success rate (left) and throughput (right) for PPO training with different observation wrapper configurations on MiniARC.}
\label{fig:ppo-obs-ablation}
\end{figure}

\textbf{Answer wrapper enables effective learning:} The \textit{Answer} configuration (2 channels: working + answer) achieves $\sim$34\% success rate at 500M steps, dramatically outperforming the baseline \textit{None} configuration (1 channel: working only) which plateaus at $\sim$6\%. Providing the target grid as an observation channel enables the policy to learn goal-conditioned behavior, i.e., it observes the desired outcome and learns action sequences to reach it. This reduces exploration burden and provides a strong learning signal aligned with the objective, though it does not reflect true generalization since the answer is given.

\textbf{Combined wrappers show diminishing returns:} The \textit{Input + Answer} configuration (3 channels) reaches $\sim$32\% success, performing comparably to Answer-only despite having the input grid available. More surprisingly, the full \textit{Input + Contextual + Answer} configuration (13 channels) achieves only $\sim$21\% success. This suggests two insights: (1) the shallow CNN architecture may lack capacity to effectively leverage high-dimensional observations, particularly the 10 contextual channels, and (2) demonstration examples introduce noise or require explicit meta-learning mechanisms (e.g., attention over demonstrations) to extract useful patterns. Simply concatenating channels does not guarantee better learning.

\textbf{Individual wrappers perform poorly:} Configurations using only \textit{Input} (2 ch, $\sim$11\%), only \textit{Contextual} (11 ch, $\sim$15\%), or \textit{Input + Contextual} (12 ch, $\sim$12\%) achieve minimal success, barely exceeding the baseline. This indicates that observing demonstrations or maintaining reference to the initial state provides little benefit without the answer grid "hint", or we need a more sophisticated architecture to extract value from these channels.

\textbf{Throughput scales inversely with channel count:} As shown in Figure~\ref{fig:ppo-obs-ablation}, the baseline None configuration (1 ch) maintains the highest throughput at approximately 1.15M steps/sec throughout training. Configurations with moderate channel counts (Answer, Input+Answer, Input) cluster around 950K-1M steps/sec. The most complex configurations show lower throughput: Contextual (11 ch) and Input+Contextual (12 ch) achieve approximately 550K steps/sec, while the full Input+Contextual+Answer configuration (13 ch) operates at approximately 500K steps/sec. Despite this 50\% reduction from baseline to full configuration, all variants maintain training throughput orders of magnitude higher than ARCLE's simulation throughput, enabling rapid experimentation across observation spaces.

\subsection{Implications for RL on ARC}

It is important to emphasize that these experiments represent a simple baseline using a shallow CNN architecture and vanilla PPO. The modest success rates (34\% on Answer configuration, 21\% on full configuration) and the apparent inability to effectively leverage contextual demonstrations suggest that significantly more sophisticated approaches will be required to tackle the full complexity of ARC reasoning tasks. The shallow CNN architecture likely lacks the representational capacity to extract and compose the abstract transformation rules that characterize ARC problems.

These results suggest a hierarchy of observation utility for policy learning on ARC tasks: answer grids provide the most value, contextual demonstrations require careful architectural design to exploit, and simply adding channels does not guarantee better performance. Future work will likely need to explore: (1) deeper architectures (e.g., ResNets, Vision Transformers) with explicit attention mechanisms to process contextual demonstrations; (2) meta-learning algorithms (e.g., MAML) specifically designed to extract transformation rules from few examples; (3) curriculum learning strategies that progressively increase task complexity; (4) hybrid neuro-symbolic approaches combining learned policies with program synthesis; (5) recurrent or memory-augmented architectures to maintain compositional state across multiple operations.

The observation wrapper system implemented in JaxARC enables rapid prototyping of such experiments. Researchers can specify arbitrary channel combinations via YAML configuration and sweep across observation spaces with minimal code changes. Furthermore, they can extend the wrapper system to implement novel observation settings as needed. This flexibility, combined with JaxARC's high performance, opens the door to large-scale empirical studies on how different inductive biases and architectures impact learning on ARC tasks.

\end{document}